\documentclass[10pt,letterpaper,twocolumn,american,letterpaper,10pt,twocolumn]{article}

\usepackage[T1]{fontenc}
\usepackage[latin9]{inputenc}
\usepackage{array}
\usepackage{amsmath}
\usepackage{amssymb}
\usepackage{graphicx}

\makeatletter

\providecommand{\tabularnewline}{\\}

\usepackage[pagenumbers]{cvpr} 

\usepackage{cellspace}
\usepackage[pagebackref,breaklinks,colorlinks]{hyperref}

\usepackage[capitalize]{cleveref}
\crefname{section}{Sec.}{Secs.}
\Crefname{section}{Section}{Sections}
\Crefname{table}{Table}{Tables}
\crefname{table}{Tab.}{Tabs.}

\def\cvprPaperID{23} 
\def\confName{CVsports}
\def\confYear{2024}
\newcommand{\mysection}[1]{\vspace{2pt}\noindent\textbf{#1}}

\makeatother

\usepackage{babel}
\begin{document}

\author{Floriane Magera$^{1,2}$
\quad
Thomas Hoyoux$^{1}$  
\quad
Olivier Barnich$^{1}$  
\quad
Marc Van Droogenbroeck$^{2}$  
\and $^1$ {\small EVS Broadcast Equipment}
\quad $^2$ {\small University of Li{\`e}ge, Belgium}
}
\inputencoding{latin9}\global\long\def\focal#1{f_{#1}}%

\global\long\def\transpositionSymbol{T}%

\global\long\def\translationVector{\mathbf{\mathbf{t}}}%

\global\long\def\time{t}%

\global\long\def\rotationMatrix{\mathbf{\text{\textbf{R}}}}%

\global\long\def\calibrationMatrix{\mathbf{\text{\textbf{K}}}}%

\global\long\def\principalPoint{p}%

\global\long\def\principalPointX{\principalPoint_{x}}%

\global\long\def\principalPointY{\principalPoint_{y}}%

\global\long\def\skew{s}%

\global\long\def\distanceToPrincipalPoint{r}%

\global\long\def\coordinateImagePlaneX{i}%

\global\long\def\coordinateImagePlaneY{j}%

\global\long\def\radialDistorsionParameterOne{k_{1}}%

\global\long\def\radialDistorsionParameterTwo{k_{2}}%

\global\long\def\class{c}%

\global\long\def\polyline{L}%

\global\long\def\estimatedPolyline{\hat{l}}%

\global\long\def\location{x}%

\global\long\def\pointOfEstimatedClass{\hat{\location}_{\class}}%

\global\long\def\threshold{\tau}%

\global\long\def\completenessRatio{\text{CR}}%

\global\long\def\time{t}%

\global\long\def\pitchWidth{W}%

\global\long\def\pitchLength{L}%

\global\long\def\pitchElevation{e}%

\global\long\def\tn{\mathrm{tn}}%

\global\long\def\fp{\mathrm{fp}}%

\global\long\def\fn{\mathrm{fn}}%

\global\long\def\tp{\mathrm{tp}}%

\global\long\def\numN{\mathsf{N}}%

\global\long\def\numP{\mathsf{P}}%

\global\long\def\numTN{\mathsf{T}\numN}%

\global\long\def\numFP{\mathsf{F}\numP}%

\global\long\def\numFN{\mathsf{F}\numN}%

\global\long\def\numTP{\mathsf{T}\numP}%

\global\long\def\proportion{\mathsf{p}}%

\global\long\def\proportionOfnumTN{\proportion\numTN}%

\global\long\def\proportionOfnumFP{\proportion\numFP}%

\global\long\def\proportionOfnumFN{\proportion\numFN}%

\global\long\def\proportionOfnumTP{\proportion\numTP}%

\global\long\def\proba#1{P\left(#1\right)}%

\global\long\def\clazz{y}%

\global\long\def\cpos{c^{+}}%

\global\long\def\cneg{c^{-}}%

\global\long\def\estimate#1{\widehat{#1}}%

\global\long\def\metric{\text{I}}%

\global\long\def\ppv{\text{PPV}}%

\global\long\def\npv{\text{NPV}}%

\global\long\def\tpr{\text{TPR}}%

\global\long\def\tnr{\text{TNR}}%

\global\long\def\fprE{\mbox{\ensuremath{\left(1-\tnr\right)}}}%

\global\long\def\fpr{\text{FPR}}%

\global\long\def\fnr{\mbox{\ensuremath{\left(1-\tpr\right)}}}%

\global\long\def\fprx{\text{FPR}}%

\global\long\def\fnrx{\text{FNR}}%

\global\long\def\accuracy{\text{A}}%

\global\long\def\errorrate{\text{ER}}%

\global\long\def\balancedaccuracy{\text{BA}}%

\global\long\def\jaccard{\text{JaC}}%

\global\long\def\auc{\text{AUC}}%

\global\long\def\bias{\text{B}}%

\global\long\def\priorSymbol{\pi}%

\global\long\def\priorneg{\priorSymbol{}^{-}}%

\global\long\def\priorpos{\priorSymbol^{+}}%

\global\long\def\rateOfNegativePredictions{\tau^{-}}%

\global\long\def\rateOfPositivePredictions{\tau^{+}}%

\global\long\def\Fscore{\text{F}_{1}}%

\global\long\def\recall{\text{R}}%

\global\long\def\precision{\text{P}}%

\global\long\def\cardinality{\#}%

\global\long\def\comma{\,,}%

\global\long\def\point{\,.}%

\global\long\def\wcdataset{\text{WC14}}%


\global\long\def\iou{\text{IoU}}%

\global\long\def\ioupart{\iou_{\text{part}}}%

\global\long\def\iouwhole{\iou_{\text{whole}}}%


\global\long\def\pitch{\text{P}}%

\global\long\def\setOfSemanticElements{\mathcal{S}}%

\global\long\def\numberOfSemanticElements{\cardinality\allClasses}%

\global\long\def\aSemanticElement{s}%

\global\long\def\aPointOfASemanticElement{p}%

\global\long\def\aPointInThreeD{X}%

\global\long\def\aPointInTheTwoDImage{x}%

\global\long\def\projectionSymbol{\pi}%

\global\long\def\projectionOfAThreePoint#1{\projectionSymbol\left(#1\right)}%

\global\long\def\calibrationMetric{\jaccard}%

\global\long\def\calibrationMetricWithThreshold#1{\calibrationMetric_{#1}}%

\newcommand{\protocolName}{ProCC\xspace}
\newcommand{\metricName}{JaC\xspace}

\title{A Universal Protocol to Benchmark Camera Calibration for Sports}
\maketitle
\begin{abstract}
Camera calibration is a crucial component in the realm of sports analytics, as it serves as the foundation to extract 3D information out of the broadcast images. Despite the significance of camera calibration research in sports analytics, progress is impeded by outdated benchmarking criteria. 
Indeed, the annotation data and evaluation metrics provided by most currently available benchmarks strongly favor and incite the development of sports field registration methods, i.e. methods estimating homographies that map the sports field plane to the image plane. However, such homography-based methods are doomed to overlook the broader capabilities of camera calibration in bridging the 3D world to the image. In particular, real-world non-planar sports field elements (such as goals, corner flags, baskets, ...) and image distortion caused by broadcast camera lenses are out of the scope of sports field registration methods.
To overcome these limitations, we designed a new benchmarking protocol, named \protocolName, based on two principles: (1) the protocol should be agnostic to the camera model chosen for a camera calibration method, and (2) the protocol should fairly evaluate camera calibration methods using the reprojection of arbitrary yet accurately known 3D objects. 
Indirectly, we also provide insights into the metric used in SoccerNet-calibration, which solely relies on image annotation data of viewed 3D objects as ground truth, thus implementing our protocol. With experiments on the World Cup 2014, CARWC, and SoccerNet datasets, we show that our benchmarking protocol provides fairer evaluations of camera calibration methods. By defining our requirements for proper benchmarking, we hope to pave the way for a new stage in camera calibration for sports applications with high accuracy standards.

\end{abstract}

\section{Introduction\label{sec:Introduction}}

\begin{figure}
\begin{centering}
\includegraphics[viewport=112.5bp 67.534bp 1440bp 712.859bp,clip,width=1\columnwidth]{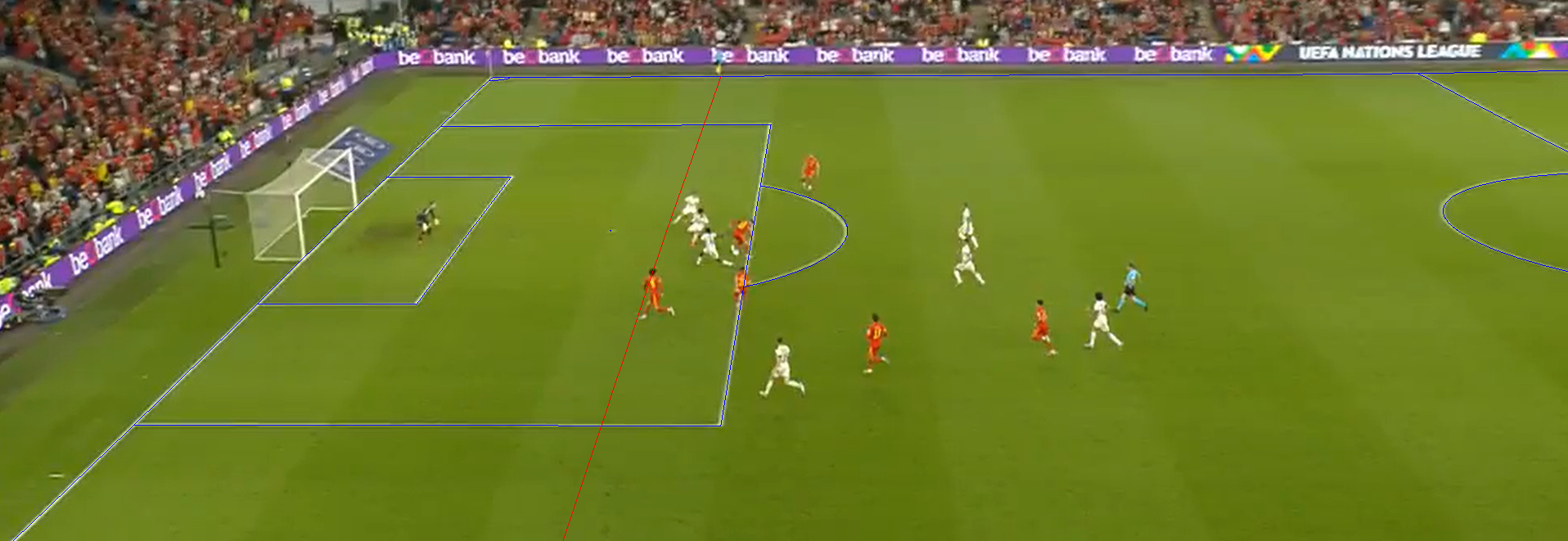}
\par\end{centering}
\caption{Illustration of a successful camera calibration. Lines superimposed
in blue are obtained by projecting the markings of a soccer field.
A perfect alignment is mandatory to enable high-precision applications
such as offside position assessment (in red, a parallel to the goal
line to decide on an offside situation). This paper proposes a new protocol, named \protocolName, that has two advantages: (1) it is applicable to any sport, and (2) its evaluation is based on a new metric which is agnostic to the chosen camera type and model.\label{fig:graphical-abstract}}
\end{figure}

Camera calibration, also known as camera resectioning, is a necessity
in many computer vision tasks. It involves estimating the parameters
of a camera model, usually the pinhole camera model, that approximates
the physical camera that produces a given image. Applications that
require accurate calibration are varied and include virtual reality~\cite{Lu2007Single},
underwater measurement~\cite{Ma2023ACombined,Shortis2019Camera},
traffic analysis and surveillance~\cite{Kikkawa2023Accuracy,Quispe2022ISee-arxiv,Tang2022Novel},
vehicle localization~\cite{Zhang2021Vehicle} or speed estimation~\cite{Schoepflin2003Dynamic},
person re-identification~\cite{Xu2021WideBaseline-arxiv},
3D reconstruction~\cite{Kano2020Accurate}, \etc. In this paper
however, we focus on calibration of cameras used in sports event broadcasting
as shown in Figure~\ref{fig:graphical-abstract}, and present a new
protocol for the evaluation of the camera calibration quality
that is usable for any camera type (static/moving, PTZ, fish-eye,
wide angle) and model (pinhole, with/without radial distortion, $\etc$).

\mysection{Camera calibration for sports}. In the context of sport,
computer vision systems are used for multiple purposes, such as augmented
reality graphics~\cite{Rematas2018Soccer}, 3D ball trajectory reconstruction~\cite{Chao20233DBall}
or tracking~\cite{Wu2020Multi-camera}, refereeing~\cite{Held2023VARS,Held2024XVARS,Kurowski2018Accurate},
or the computation of statistics regarding position, speed, and acceleration
of balls, players, bats, sticks, pucks, \etc~\cite{Maglo2023Individual,VanZandycke20223DBall,Jiang2022GolfPose,Vats2021Puck}.
The expectations in terms of precision and robustness of these systems
is constantly rising. Both the fact that it is considered as a valid
alternative to GPS trackers for player tracking systems and the fact
that the FIFA supported the use of the SAOT (Semi-Automated Offside
Technology) for the 2022 Football World Cup are good indicators of
the trust put in computer vision systems. Sports analysis is an active
subject of research, with several recent datasets gathering broadcast
images with a wide range of annotations serving for game analysis,
such as DeepSportRadar dataset for Basketball~\cite{VanZandycke2022DeepSportradarv1},
or the SoccerNet datasets for soccer \cite{Cioppa2022SoccerNetTracking,Deliege2021SoccerNetv2,Giancola2018SoccerNet, Leduc2024SoccerNetDepth}.
The SoccerNet dataset which was initially created for action spotting tasks \cite{Giancola2018SoccerNet,Cabado2024Beyond} covers a wide range of tasks for soccer analysis.
In particular, the game state recognition task, one
of the latest challenges proposed by the SoccerNet team, heavily relies on the ability to calibrate the camera to produce a strategic
minimap of the players' localization~\cite{Somers2024SoccerNetGameState}.
This challenge paves the way to a 3D reconstruction that will require
to be able to calibrate any camera along the field, including close-up
cameras in a multi-view setup. 

Camera calibration in the context of sports benefits from the presence
of the sports field whose dimensions are specified by the rules of
the game (see for example~\cite[page 32]{IFAB2022Laws} for soccer,
\cite{FIBA2022Official} for basketball, \cite{FIVB2021Official} for volleyball, and~\cite{IIHF2022Official} for ice hockey). Their
well-known shape and their presence in most sports images make them
convenient calibration patterns. Using the field model as a calibration
pattern, camera calibration techniques can thus express the parameters
by a projection matrix, and sports field registration techniques express
the mapping between the field plane and the image by a perspective
transform, better known as a homography~\cite{Szeliski2022Computer}. Note, however, that the rules of games allow some
tolerance on field dimensions, so calibration systems have to cope
with some dimension uncertainties. In this regard, goal posts are
supposed to be perfectly sized, which makes them adequate to serve
as calibration landmarks.

As the terms of ``camera'' calibration and ``sports field'' registration can
be sometimes used interchangeably in the literature, we deem important
to emphasize the difference between them. A sports field registration
technique aims to estimate a homography, \ie the transformation between
the 3D sports field plane and the image. This mapping is not defined
outside these two planes, which is insufficient for all 3D applications
that involve elements outside the field plane. This is a blind spot
in all papers that mention 3D applications such as player and ball
tracking, tactics analysis, augmented reality, \etc. On the other
hand, a camera calibration technique provides a mapping between the
3D world (not only the field plane) and the image, which makes it
suited for the aforementioned applications. Sports field registration
is an approximate attempt or a first step towards camera calibration,
and this is why, in this paper, we consider the homography as a camera
model, despite its practical limitations. However, we show that our
protocol is applicable to any model, broadening the possibilities
in terms of camera models.

\mysection{Outline.} The rest of the paper is organized as follows.
We present the related work on camera calibration for sports, including
a review of existing datasets, in Section~\ref{sec:Related-work}.
Section~\ref{sec:Methods} describes the key elements of our benchmarking
protocol, named \protocolName, the new metric 
\metricName\footnote{We have renamed the metric from AC, used in SoccerNet-calibration, to \metricName to avoid any confusion with the notion of accuracy as used in binary classification.} 
that we introduced in SoccerNet-calibration, as well as a more relevant ground-truth type, to express the camera calibration quality. 
Our benchmarking protocol, including the annotation procedure and its metric, are the main contributions of this paper. Section~\ref{sec:Results} applies our protocol to the
evaluation of camera calibration for sports broadcast events. This
section also includes an experimental analysis of the protocol for
different camera models, illustrating the universality of our protocol
and showing that previous protocols have limitations that our new
protocol can overcome. Section~\ref{sec:Discussion} discusses the
results and explains why current calibration techniques developed in the literature overlook the broader capabilities of camera calibration in bridging the 3D world to the image. Finally, Section~\ref{sec:Conclusion} presents a brief
conclusion.

\section{Related work\label{sec:Related-work}}

In the current state of the art, most of the evaluation procedures
are based on homography annotations (to the best of our knowledge,
basketball~\cite{Istasse2023DeepSportradarv2,Huang2022Pose2UV} and
athletics~\cite{Baumgartner2023Monocular} are the only exceptions)
which restricts the evaluation to the sports field plane, even if
some techniques compute camera parameters. Indeed, all the following
camera calibration techniques in the literature~\cite{Chen2019Sports,Sha2020EndtoEnd,Citraro2020Realtime,Theiner2023TVCalib}
use the pinhole model, whose camera parameters can be converted to
the homography that maps the image to the sports field plane.

\subsection{Datasets to benchmark camera calibration}

The latest techniques in camera calibration for sports
all leverage the presence of a sports field to understand the mapping
between the image and the world.

By browsing the scientific literature, we found a dozen datasets on
which researchers evaluated their methods. Table~\ref{tab:Current-datasets}
shows that most datasets are only relevant for sports field registration,
as their annotations consist of homographies. For some datasets, it
seems that the ground truth has not been annotated once and for all,
such that several researchers have created their own annotations,
which does not ensure the comparability of their results~\cite{Shi2022Self}.

The nature of the annotations, the relatively small sizes of the datasets
for some sports, as well as the difficulty of getting access to the
data, make the creation of reliable camera calibration algorithms
difficult.

\begin{table*}
\begin{centering}
\begin{tabular}{lccccc}
\hline 
{Name} & Sport & Open data & Size & Annotations & Techniques\tabularnewline
\hline 
WorldCup 14 & Soccer & Yes & $395$ & Homographies & \cite{Chen2019Sports,Sha2020EndtoEnd,Citraro2020Realtime,Theiner2023TVCalib,Shi2022Self,Homayounfar2017Sports,Sharma2018Automated,Jiang2020Optimizing,Nie2021Robust,Chu2022Sports,Zhang2022AFast,Cioppa2021Camera,Li2021Soccer,Tsurusaki2021Sports,Maglo2023Individual}\tabularnewline
TS-WorldCup & Soccer & Yes & $3{,}812$ & Homographies & \cite{Chu2022Sports}\tabularnewline
SoccerNet-calibration & Soccer & Yes & $21{,}132$ & Field markings & \cite{Theiner2023TVCalib}\tabularnewline
CARWC & Soccer & Yes & $4{,}207$ & Homographies & \cite{Claasen2023Video}\tabularnewline
{SportLogiq} & Ice Hockey & No & $1.67$M & Not specified & \cite{Shi2022Self,Homayounfar2017Sports,Jiang2020Optimizing}\tabularnewline
SportsFields by Amazon & Multi-sports & No & $2{,}967$ & Homographies & \cite{Nie2021Robust}\tabularnewline
{Volley ball} & Volley ball & Yes & 470 & Homographies & \cite{Chen2019Sports,Shi2022Self}\tabularnewline
{College Basketball}& Basketball & No & 640 & Homographies & \cite{Sha2020EndtoEnd}\tabularnewline
DeepSportRadar & Basketball & Yes & 728 & Pinhole model & \cite{VanZandycke2022DeepSportradarv1}\tabularnewline
3DMPB & Basketball & Yes & 10k & Pinhole model & \cite{Huang2022Pose2UV,Shu2022ACamera}\tabularnewline
{Athletics}& Athletics & Yes & 10k & Pinhole model & \cite{Baumgartner2023Monocular}\tabularnewline
\end{tabular}
\par\end{centering}

\caption{Main current datasets dedicated to camera calibration
for different sports. In this table, we also mention if the annotations are available, the number of images (Size) for which annotations are provided, if the annotations are specific to homographies or the pinhole model, and list some camera calibration techniques using a given dataset. From this table, one can see that the WorldCup 14 dataset is the most popular dataset for benchmarking, despite its limitations, as explained in this paper.}\label{tab:Current-datasets}
\end{table*}

\subsection{Evaluation of camera calibration in sports}

\begin{figure}
\begin{centering}
\includegraphics[width=\columnwidth]{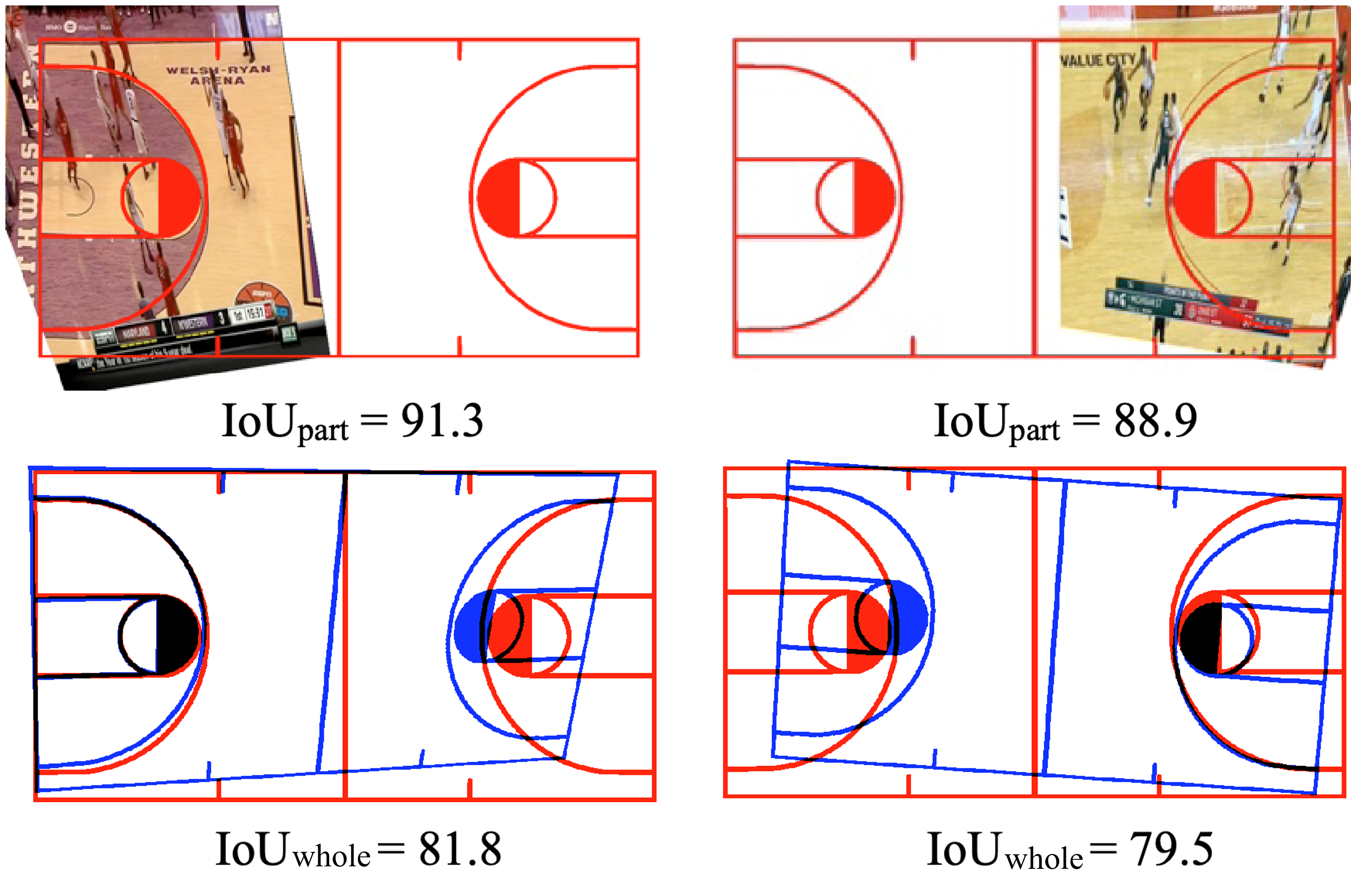}
\par\end{centering}
\caption{Visualization of the $\protect\iouwhole$ and $\protect\ioupart$
metrics. Illustration taken from~\cite{Sha2020EndtoEnd} (\textcopyright IEEE, 2020). \label{fig:Visualization-of-iou-metric}}
\end{figure}

Today, most authors evaluate their performance using an intersection over union metric, more specifically the $\iouwhole$
metric, that was first introduced by Homayounfar \etal~\cite{Homayounfar2017Sports}
for the performance evaluation of their algorithm on soccer and ice
hockey content. This metric measures the intersection over union between
the real-world rectangle defined by the field template and its successive
projection then deprojection using both the annotated homography and
the camera parameters in the 3D world. This metric was extended by
Sharma \etal~\cite{Sharma2018Automated} into another
metric, denoted by $\ioupart$, to only account for the parts of the
field that are visible in the image. These two metrics, which are
illustrated in Figure~\ref{fig:Visualization-of-iou-metric}, were
applied to other sports such as volleyball~\cite{Chen2019Sports,Shi2022Self},
basketball~\cite{Sha2020EndtoEnd}, tennis and American football~\cite{Nie2021Robust}.

In 2022, we launched the SoccerNet-calibration challenge, which is an attempt to further improve
calibration techniques, but despite the organization of two SoccerNet-calibration
challenges~\cite{Giancola2022SoccerNet,Cioppa2023SoccerNetChallenge-arxiv},
the benchmark of the World Cup 14 (WC14) dataset remains popular and the
most commonly used benchmark. Therefore, we have collected the results
of the current state-of-the-art methods, organized by their year of publication,  regarding the WC14 benchmark
in a consolidated leaderboard given in Table~\ref{tab:WorldCup14-leaderboard}. However, the  evaluation of one method stands out: Theiner \etal\cite{Theiner2023TVCalib} evaluated their method both on the WC14 and SoccerNet-calibration datasets and, even further, they already showcased some discrepancies between our protocol \protocolName and the WC14 evaluation protocol when applied to the same dataset.  

\begin{table}
\resizebox{\linewidth}{!}{
\begin{tabular}{ccccc}
\hline 
\noalign{\vskip0.1cm}
Reference & $\text{mIoU}_{\text{whole}}$ & $\text{medIoU}_{\text{whole}}$ & $\text{mIoU}_{\text{part}}$ & $\text{medIoU}_{\text{part}}$\tabularnewline[0.1cm]
\hline 
Homayounfar \etal~\cite{Homayounfar2017Sports} & 83 & - & - & -\tabularnewline
Sharma \etal~\cite{Sharma2018Automated} & - & - & 91.4 & 92.7\tabularnewline
Chen and Little~\cite{Chen2019Sports} & 89.2 & 91.0 & 94.7 & 96.2\tabularnewline
Jiang \etal~\cite{Jiang2020Optimizing} & 89.8 & 92.9 & 95.1 & 96.7\tabularnewline
Sha \etal~\cite{Sha2020EndtoEnd} & 88.3 & 92.1 & 93.2 & 96.1\tabularnewline
Citraro \etal~\cite{Citraro2020Realtime} & 90.5 & 91.8 & - & -\tabularnewline
Cioppa \etal~\cite{Cioppa2021Camera} & 79.8 & 81.7 & 88.5 & 92.3\tabularnewline
Li \etal~\cite{Li2021Soccer} & 92.1 & 94.3 & 95.1 & 96.7\tabularnewline
Tsurusaki \etal~\cite{Tsurusaki2021Sports} & - & - & \textbf{97} & -\tabularnewline
Nie \etal~\cite{Nie2021Robust} & 91.6 & 93.4 & 95.9 & 97.1\tabularnewline
Shi \etal~\cite{Shi2022Self} & \textbf{93.2} & \textbf{94.9} & 96.6 & \textbf{97.8}\tabularnewline
Chu \etal~\cite{Chu2022Sports} & 91.2 & 93.1 & 96.0 & 97.0\tabularnewline
Zhang and Izquierdo~\cite{Zhang2022AFast} & 90.0 & 92.8 & 95.3 & 96.9\tabularnewline
Theiner and Ewerth~\cite{Theiner2023TVCalib} & - & - & 95.3 & 96.6\tabularnewline[0.1cm]
Maglo \etal~\cite{Maglo2023Individual} & 92 & 94.1 & 96.3 & 97.4\tabularnewline[0.1cm]
\hline 
\end{tabular}}

\caption{Consolidated leaderboard by collecting published values
on the World Cup 14 benchmarking dataset. The values are the mean
or median intersections over union (denoted respectively by $\text{mIoU}$
and $\text{medIoU}$), on the whole field or on the visible part of
it, as mentioned in the references. In this table, references are
organized by their year of publication, and the best values are given
in bold. \label{tab:WorldCup14-leaderboard}}
\end{table}

Furthermore, Nie \etal~\cite{Nie2021Robust}
and Chu \etal~\cite{Chu2022Sports} provide an exhaustive evaluation by adding reprojection and projection
errors. The projection error measures the average distance in meters
between the projection of pixels sampled in the field image, using
the inversion of both the estimated homography and the ground-truth
homography. The reprojection error measures the average distance between
reprojected points in the image using the ground-truth homography
and the predicted homography, this distance is then normalized by
the image height.

All the aforementioned metrics rely on the annotated homographies
of the dataset, which are a limited and simplified interpretation
of the observable field markings in the image. Moreover, with the
small size of the World Cup 14 dataset and the small improvements
in performance obtained in the last few years, concerns have been
raised about the relevance of this dataset. Indeed, Chu \etal~\cite{Chu2022Sports}
proposed a new dataset called TS-WorldCup to increase the dataset size and allow for tracking evaluation, Claasen \etal~\cite{Claasen2023Video}
further corrected both the World Cup 14 and the TS-WorldCup annotations in a revised version named CARWC, which proposes homographies that are more precise. While the latter two improvements address some limitations
of the current benchmarking protocol, we wish to go one step further by proposing to get rid of homographies as ground-truth data for camera calibration in sports. 
In our opinion, one major problem with the current benchmarks is the imperative of a restricted and limited camera model for fulfilling the requirements of professional use. Indeed, around the field for a top-tier sports game, there can be up to 50 cameras, of a wide variety of quality: from wide-angle to super slow-motion cameras or fish-eye cameras, one model does not fit all of them. To circumvent this issue, we propose a new benchmarking protocol that allows for the use of any camera model. Furthermore, if we were to actually choose only one camera model for a set of broadcast cameras, our protocol would allow deducing which model would be the
best fit according to our model-agnostic metric ---or usual metrics such as the reprojection error--- instead of choosing an axiomatic, arbitrary camera model from the outset.

\section{A benchmarking protocol for model-agnostic camera calibration\label{sec:Methods}}

In this section, we describe our new benchmarking protocol that is
based on two main pillars: annotations and a metric. Both are designed
by assuming that a good camera calibration algorithm will be able
to produce results that allow a minimal reprojection error, which
is, in fact, the only reasonable assumption when there is no actual
ground-truth knowledge about the broadcast cameras that captured the
images of the datasets.

\subsection{Annotations: beyond homographies}

As an essential requirement for evaluating camera calibration is to
handle different camera models, we must first change the type of the
annotations and, subsequently, provide a revised evaluation metric.

In contrast to camera calibration techniques that rely on a specific
pattern that will only be seen once and self-calibration techniques
that do not require any calibration target and instead usually rely
on reference frames, sports images have the unique advantage of always
displaying at least partially a specific pattern, which is the sports
field. We suggest leveraging this advantage, which is already done
in most benchmarking and calibration techniques, but also pushing
forward, by using as ground truth a lower-level interpretation of
the sports field image: semantic point annotations of the sports field
markings, which are annotations that are valid for any type of camera.

\begin{figure}
\begin{centering}
\includegraphics[width=1\columnwidth]{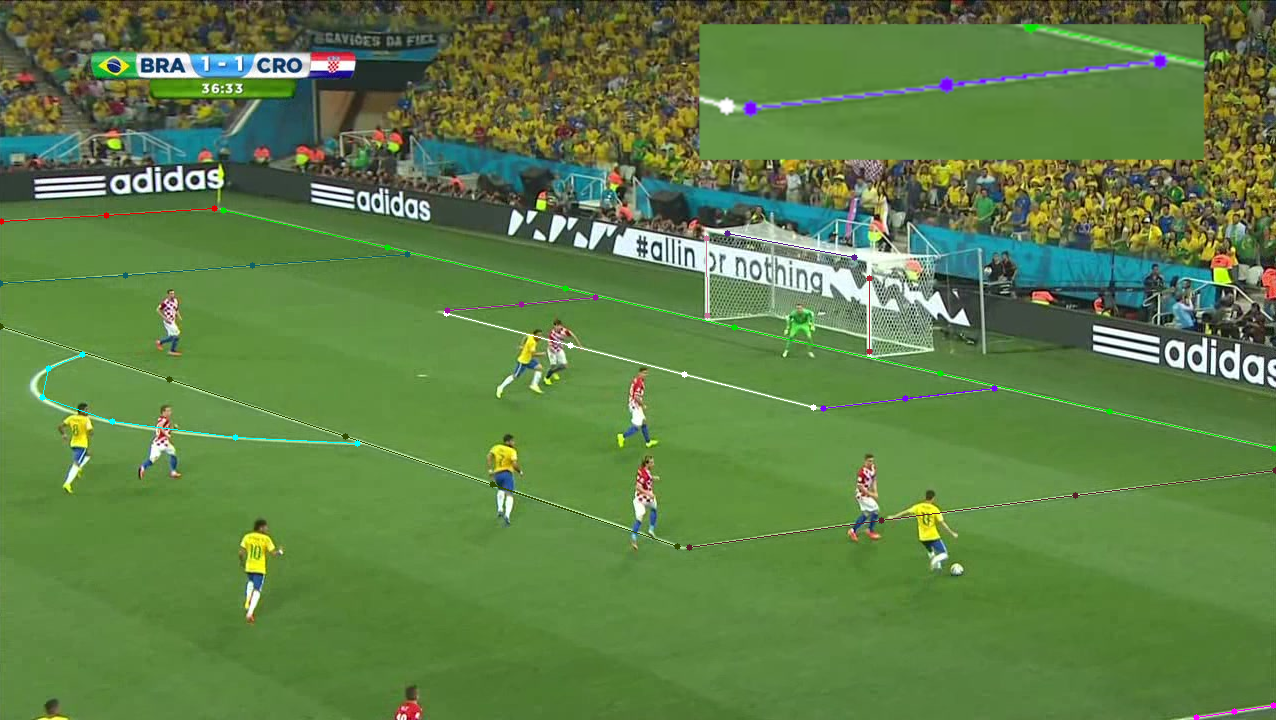}
\par\end{centering}
\caption{Illustration of annotations on a SoccerNet-v3~\cite{Cioppa2022Scaling}
image as used for the camera calibration challenge. As shown in the zoomed snapshot, annotations consist of points and the labels of the objects they belong to.\label{fig:SoccerNet-v3-annotations}}
\end{figure}

A sports field can be decomposed into a set of simple geometrical
elements that are points, lines, circles, or ellipses. For each of
these simple constituent elements of the field, we propose to assign
a unique semantic label. In general, the field is only partially visible
in the image, so for a given image, only a few elements need to be
annotated. Practically, we annotate points along each element and
thus, obtain a set of points for each semantic element of the field.
To illustrate this, in SoccerNet, the annotations consist of polylines,
\ie sequential lists of 2D points along each soccer field element.
Ideally, annotations should be regularly spaced, and the density of
annotated points might be increased depending on the curvature of
the field element as it appears in the image. Each element of the
soccer field corresponds to a class, and in total, we count $26$
semantic classes. An example of annotation is shown in Figure~\ref{fig:SoccerNet-v3-annotations};
in this figure, the points are provided by  human
annotators, while  polylines are superimposed by an annotation tool.

By employing a semantic annotation of soccer field elements represented
by a polyline, we meet the first requirement of a universal evaluation
protocol, in the sense that the ground truth is independent of the
used camera model. The main advantages of this annotation type in
the image domain are its independence from the task to be solved and
the ease to validate it by the human eye. As the annotations do not
necessarily cover all field markings pixels, we will show that our
metric addresses the issue that while the annotated points are part
of field elements, it is possible that neighboring pixels also belong to the same element, despite being unannotated.

Finally, the use of semantic annotations opens up
possibilities for more complex calibration scenarios. For example,
if a system can triangulate a 3D point and knows its projection in
the image, it may use it for calibration, even if the point belongs to a moving object such as a player.
This opens the way to close-up camera calibration in a multi-view
setup. 

We now show how to evaluate  a camera calibration
based on this ground truth.

\subsection{Evaluation metric}

Given a field model and the corresponding annotations in the image,
we can evaluate the quality of camera parameters by assessing how well the camera projection of the 3D field superimposes on the annotations.
A camera model definition includes the mathematical imaging function
that allows for the computation of the 2D image of a 3D object. 
This function is also called the  \emph{projection} function. In our
protocol, since the evaluation must be independent of the model, the
only requirement for a camera model is that this function must be
defined and, when it is given a 3D point $\boldsymbol{\aPointInThreeD}$,
it must be able to provide its corresponding 2D location $\boldsymbol{\aPointInTheTwoDImage}$
in the image plane. We denote the projection function of a camera
model by $\projectionSymbol$ and, therefore, $\boldsymbol{\aPointInTheTwoDImage}=\projectionOfAThreePoint{\boldsymbol{\aPointInThreeD}}$.

The metric that we propose is inspired by the reprojection error,
while aiming to address its shortcomings. The reprojection error usually
measures the Euclidean distance between the observation (\eg the
annotation) of an object and the projection of that 3D object model
using the estimated camera model. The mechanism of projection is the
source of several difficulties. First, depending on the chosen camera
model, the reprojection error may be undefined when the projection
of the 3D object does not land in the image and, more generally, is
unsuited to handle hallucinated or missing objects. Second, a projection
may not exactly match an annotation, but may still be superimposed
on visible field marks, since the field markings have a thickness
that can make them visible for more than one pixel in width. Finally,
it is inconvenient to grasp the meaning of an unbounded metric that,
in our case, can vary between zero and infinity. For these reasons,
by thresholding the reprojection errors, we obtain another metric
that can be intuitively understood as the proportion of field elements
that are correctly imaged, bridging the camera calibration evaluation
with an object detection problem. Indeed, the underlying intuition
behind our metric is that the quality of the camera parameters reflects
how well the parameters can reproject objects close to their
image.

Since we defined that each field marking element is annotated with
as many points as necessary to constitute a fitting polyline, we mark
a field element as correctly detected if its reprojection in the image
is close enough to each of the annotated points of the polyline. The
closeness criterion is defined based on a threshold value $\tau$
in pixels. More formally, a field element is defined by a polyline
$L$, which is an ordered list of 3D points. Its projection $\pi(L)$
gives a 2D polyline (\ie an ordered list of 2D points) which also
defines a list of line segments $S$ by considering pairs of the consecutive
2D points. The distance between an annotated point $\boldsymbol{x}_{i}$
and the projection of the field element $\pi(L)$ is given by the
minimal Euclidean distance between the point $\boldsymbol{x}_{i}$
and the segments $S$ constituting $\pi(L)$.

In the following, we define the function $d(.)$ computing the distance
between a point and a line segment. More specifically,
given the projection $\boldsymbol{c}$ of the point $\boldsymbol{x}$
on the line passing through the segment extremities $\boldsymbol{a}$
and $\boldsymbol{b}$, we define the distance between the point $\boldsymbol{x}$
and the line segment $S_{\boldsymbol{ab}}$ defined by $\boldsymbol{a}$
and $\boldsymbol{b}$ with the traditional Euclidean distance function
as follows:

\begin{equation}
d(\boldsymbol{x},S_{\boldsymbol{ab}})=\begin{cases}
\left\Vert \boldsymbol{x}-\boldsymbol{a}\right\Vert _{2} & z\leq0\comma\\
\left\Vert \boldsymbol{x}-\boldsymbol{c}\right\Vert _{2} & 0<z<1\\
\left\Vert \boldsymbol{x}-\boldsymbol{b}\right\Vert _{2} & z\geq1\comma
\end{cases},z=\frac{\boldsymbol{c-}\boldsymbol{a}}{\boldsymbol{b-}\boldsymbol{a}}\comma\label{eq:dist-segment}
\end{equation}
where the variable $z$ is introduced to account for the fact that the distance to the line segment is equal to the distance to one of its extremities if the point projection $c$ lands outside the segment.
Finally, we say that a field element $L$ is correct if all the points
$\boldsymbol{x}_{i}$ annotated for this element are less than $\tau$
pixels away from the projection of said field element $L$. In mathematical
terms, we then have that:

\begin{equation}
\min_{S\in\pi(L)}d(\boldsymbol{x}_{i},S)<\tau\comma\forall\boldsymbol{x}_{i}\point\label{eq:seuil-tp}
\end{equation}

The correctly detected field markings are counted as true positives
$\numTP{}_{\tau}$. Both the hallucinated field markings and wrongly
detected field elements whose reprojection lands further than $\tau$
pixels from the annotations are false positives $\numFP$. Missing
field elements are counted as false negatives $\numFN$. Finally,
we define our $\calibrationMetricWithThreshold{\tau}$ metric, the Jaccard index for camera calibration or \metricName, to evaluate the calibration ``accuracy'' as
follows: 

\begin{equation}
\calibrationMetricWithThreshold{\tau}=\dfrac{\numTP{}_{\tau}}{\numTP{}_{\tau}+\numFN+\numFP}\point\label{eq:accuracytau}
\end{equation}
 We propose to use this metric, parametrized by the reprojection error
$\tau$, for the best comprehension of the camera parameters quality.

\section{Results\label{sec:Results}}

For practical reasons, our experiments focus on soccer because this sport has the most available techniques and
data.  Furthermore, we are restricted to soccer due to a lack of semantic annotations for other sports, despite that \protocolName is applicable to any sport for which there are measurable 3D references.
Hereafter, we describe experiments on  the WC14, CARWC,
and the SoccerNet datasets to demonstrate our benchmarking protocol.

\subsection{Description of the datasets}

For all the datasets, the experiments employ soccer field markings
and goal posts that are decomposed into distinct classes.

\mysection{SoccerNet.} The SoccerNet dataset contains $21{,}132$ images that were recently
annotated with soccer field markings elements, as well as goal posts
elements, which provides 3D correspondences, thus outside the field
plane~\cite{Cioppa2022Scaling}. In total, the annotations consist
of $167{,}589$ field marking lines or circles and $53{,}577$ goal
posts elements that are grouped into 26 classes for soccer fields.
Each element is annotated with its two extremities for rectilinear
elements and by as many points as needed to fit curves for non-rectilinear
elements.

\mysection{World Cup 14.} For the sake of comparison, the World Cup 14 test set has been manually
annotated following our convention. Indeed, we annotated the 186 images
with points along soccer field elements, resulting in a total of $1{,}681$
soccer field elements annotated with several points.

\subsection{Establishing a better camera model for broadcast cameras}

The goal of our experiments is twofold: validate our evaluation protocol, and demonstrate that there is a better camera model for broadcast cameras. 
This is why we  first compare both qualitatively and quantitatively ground-truth homographies of the WC14 and CARWC datasets with estimated camera parameters following a richer camera model.

In our experiments, we selected a threshold $\tau$ of 5 pixels for our $\calibrationMetricWithThreshold 5$ metric, which is a value that is suited for distinguishing between methods given the current quality standards of the different camera calibration methods that exist in the literature. 
When we want to differentiate quite precise methods, we tune the pixel threshold to 2 pixels.

\begin{figure*}
\begin{centering}
\begin{tabular}{ccc}
\includegraphics[width=0.31\textwidth]{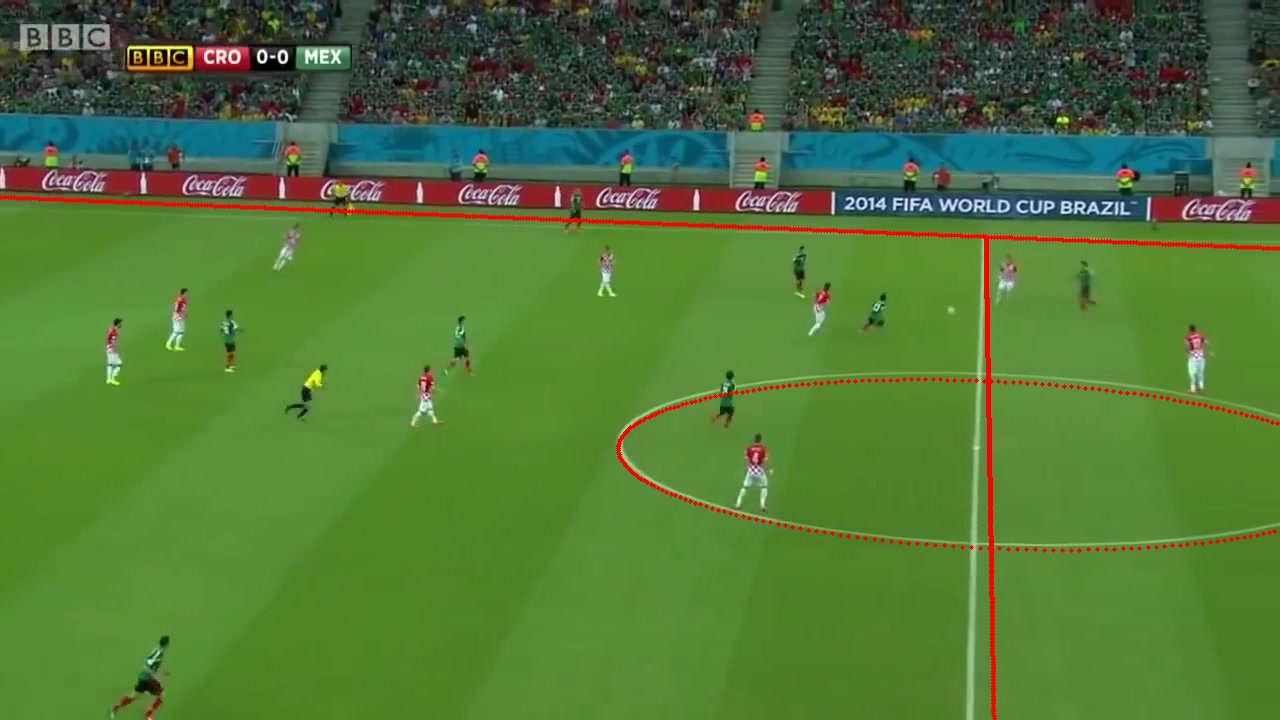} & \includegraphics[width=0.31\textwidth]{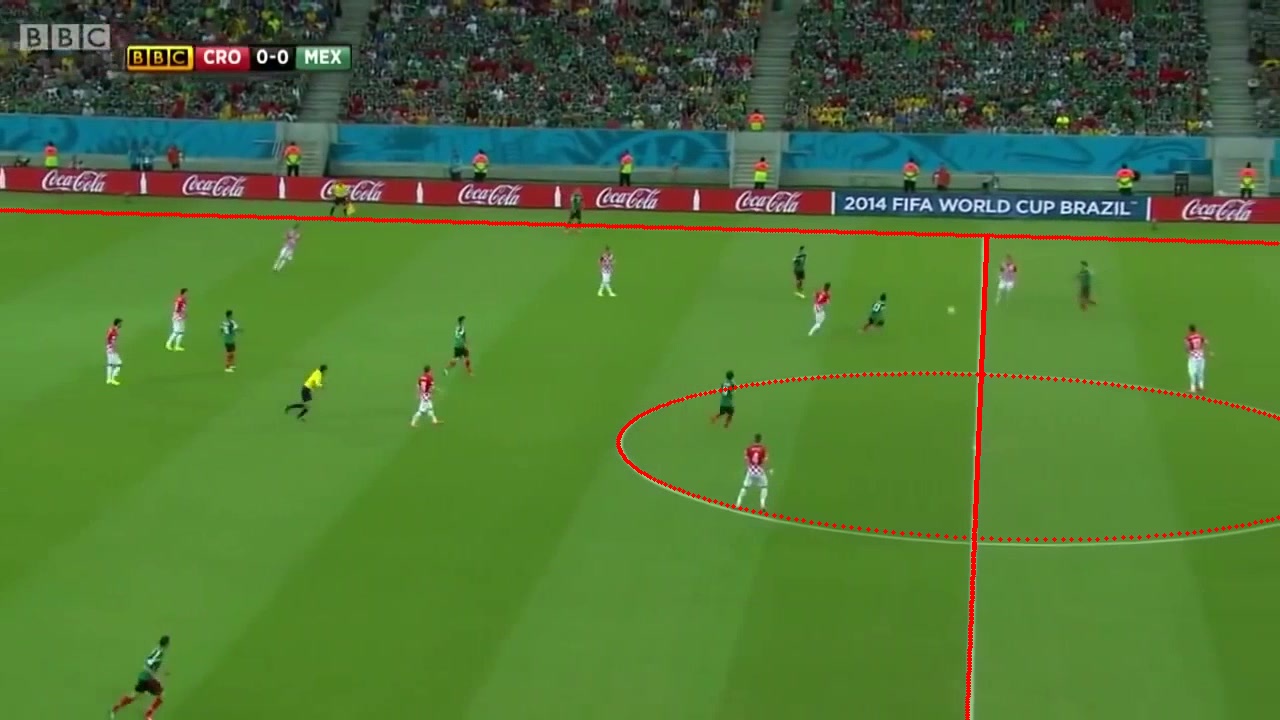} & \includegraphics[width=0.31\textwidth]{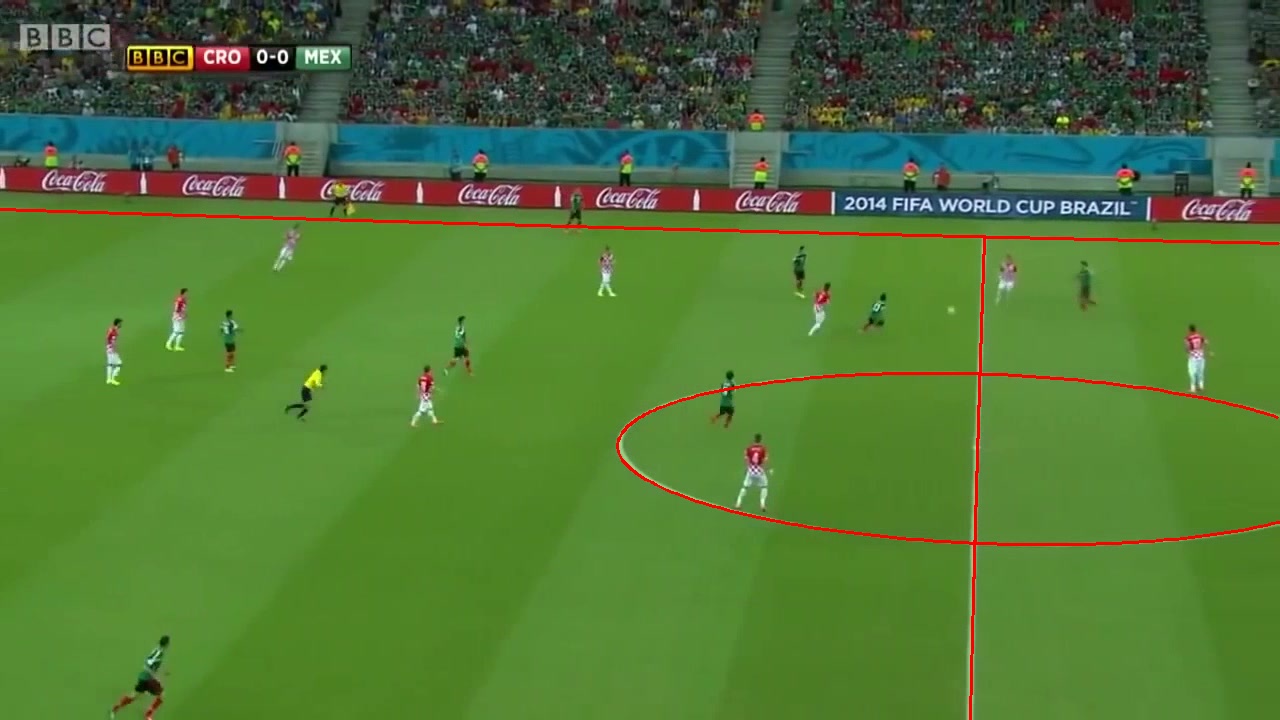}\tabularnewline
\includegraphics[width=0.31\textwidth]{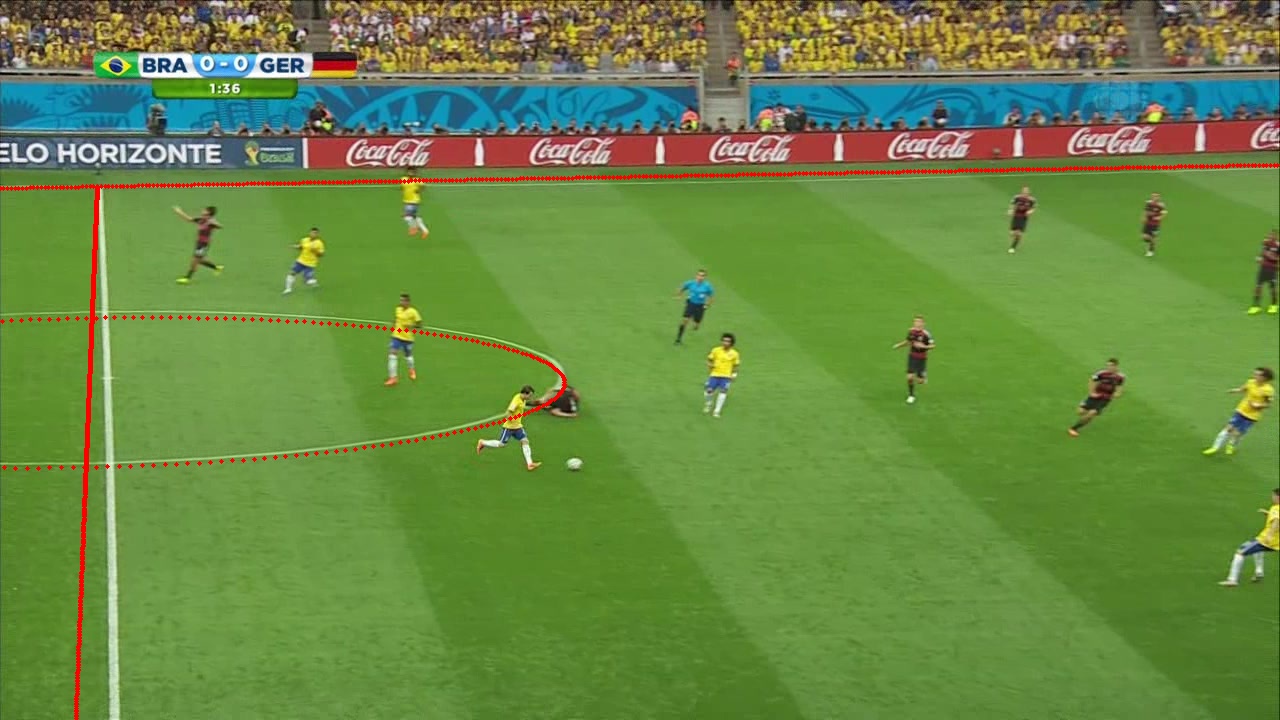} & \includegraphics[width=0.31\textwidth]{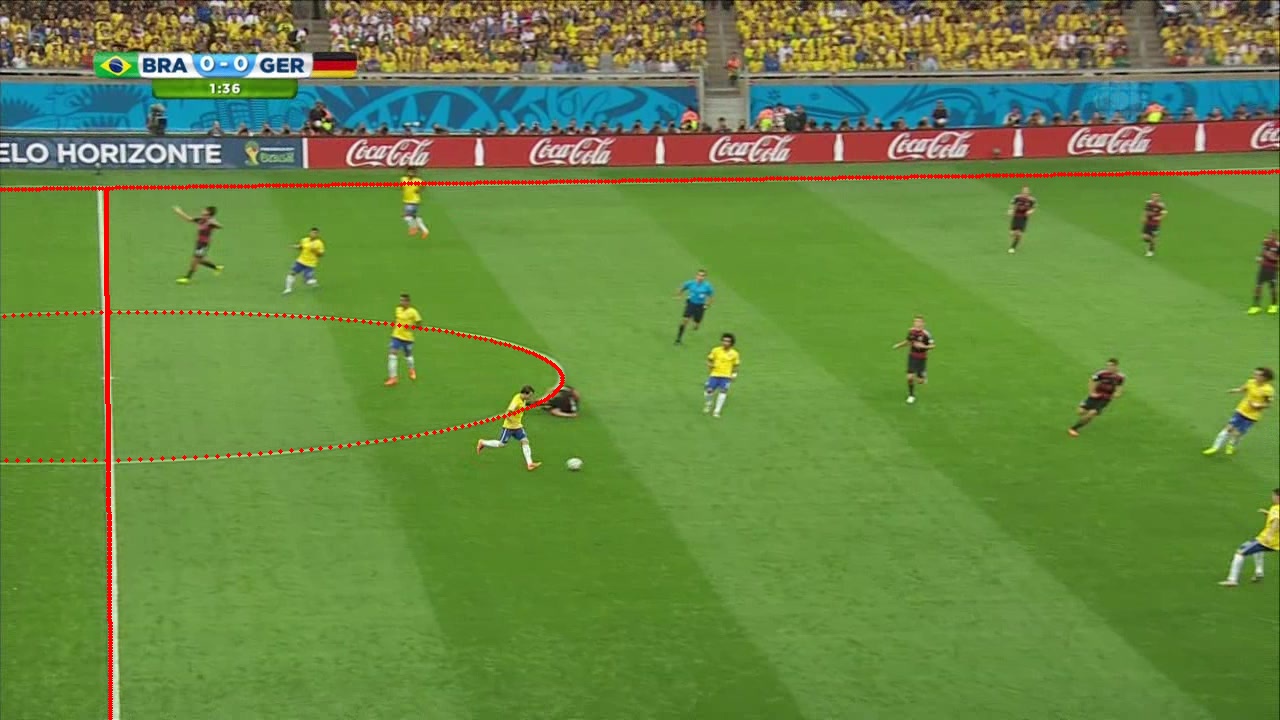} & \includegraphics[width=0.31\textwidth]{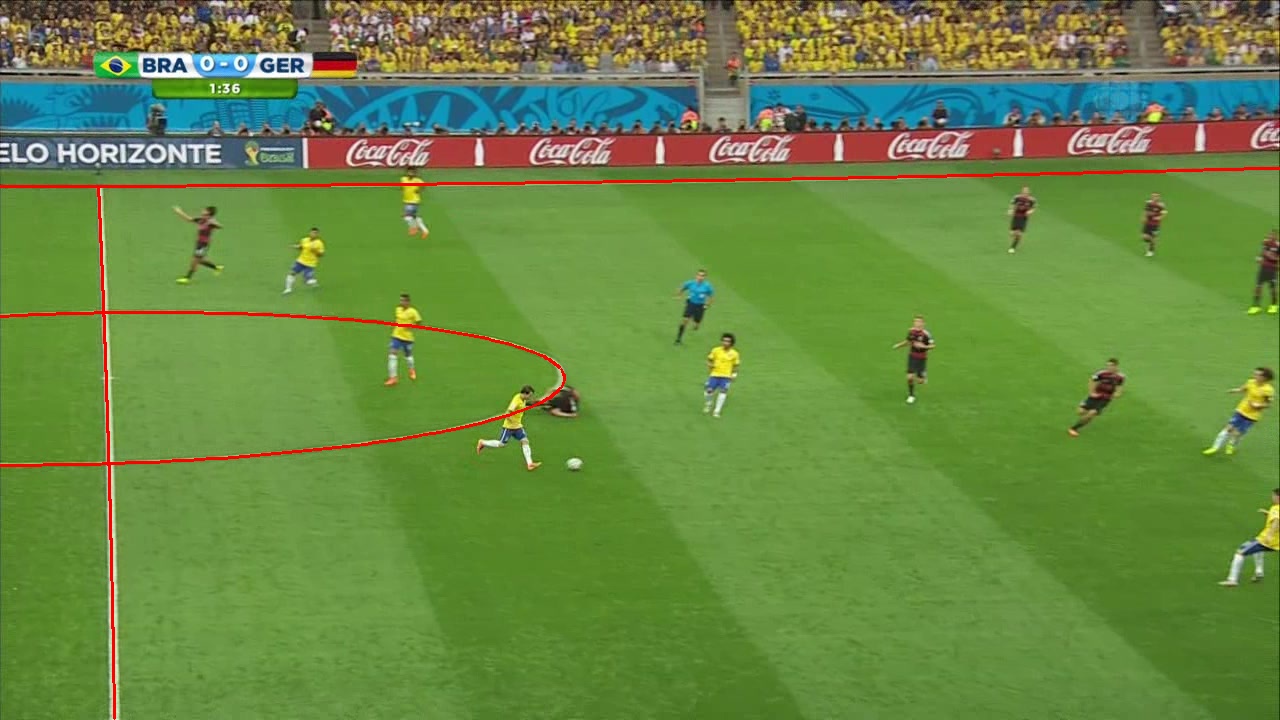}\tabularnewline
\includegraphics[width=0.31\textwidth]{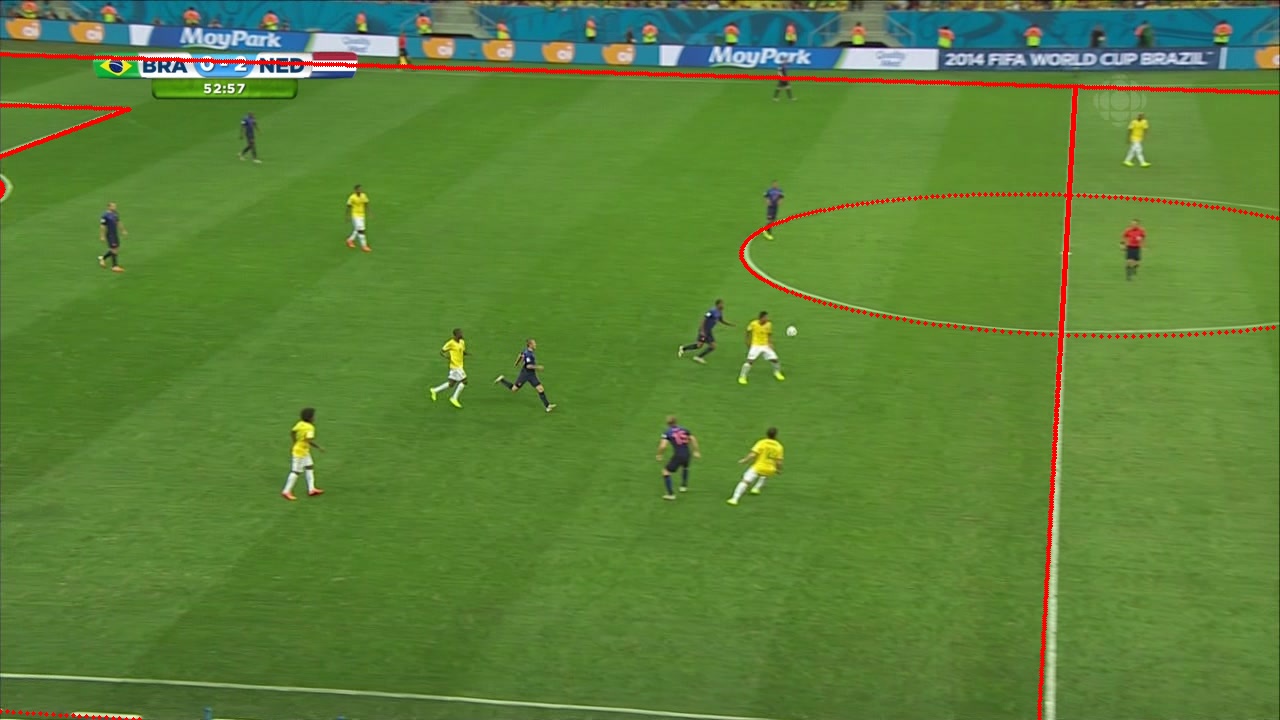} & \includegraphics[width=0.31\textwidth]{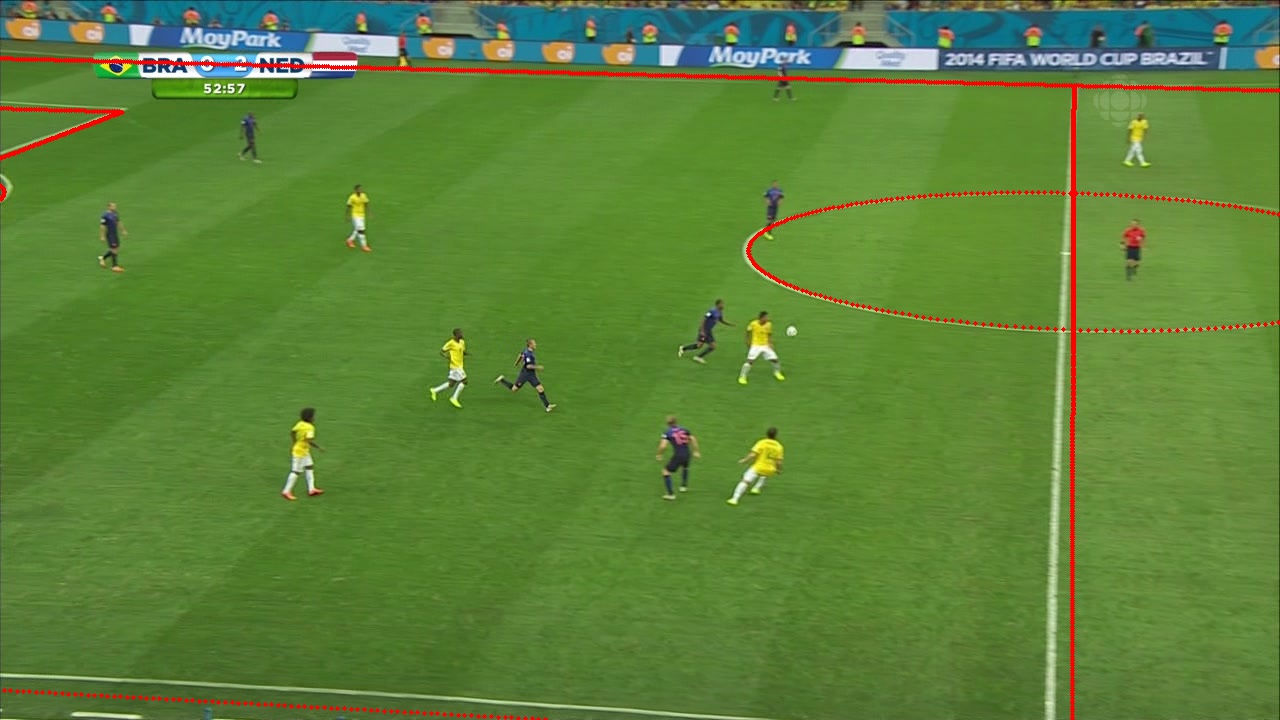} & \includegraphics[width=0.31\textwidth]{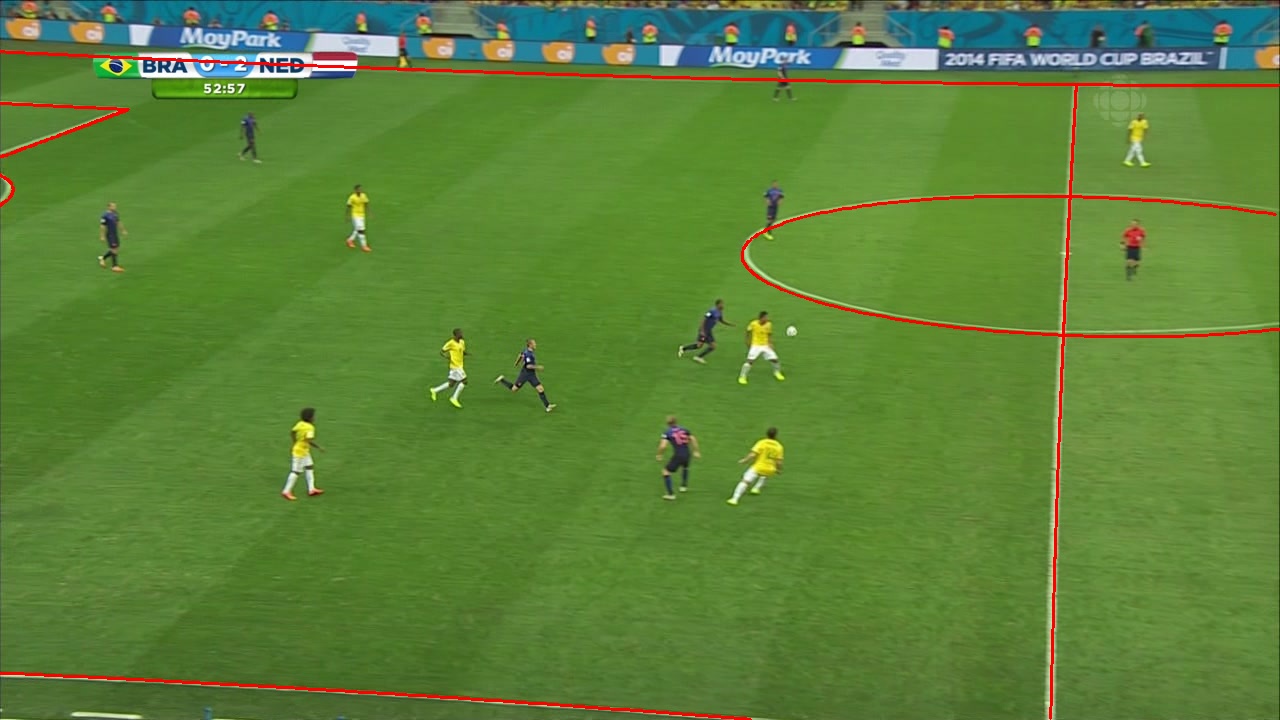}\tabularnewline
 WC14 annotations &  CARWC annotations & Xeebra \tabularnewline
\end{tabular}
\par\end{centering}
\caption{Comparison of reprojected field elements based on different types
of annotations: in the first column, the WC14 homographies are used
to project the soccer field model. The second column corresponds to
the CARWC annotations. Both these annotations fail to provide correct
reprojections. Finally, the third column displays results obtained
with the Xeebra product. \label{fig:qualitative-assessment}}
\end{figure*}

As illustrated in Figure~\ref{fig:qualitative-assessment} and explained
in the above, the reprojection of the soccer field model using \emph{ground-truth
homographies} (see columns 1 and 2) fails to match the images of the
field markings. This is an indication that the use of these annotations
is suboptimal if we agree that a good camera calibration algorithm should be able to match the image of known 3D objects such as soccer
fields, which is a reasonable and tractable solution in the absence
of actual ground truth for the camera parameters. 

In Table~\ref{tab:Comparison-of-annotations}, we compare different
camera models and highlight the limitations of using homographies
as ground truth. The manual annotations and $\calibrationMetricWithThreshold 5$
metric allow us to quantify the quality of the annotations of the
WC14 and CARWC datasets compared to the camera parameters estimated
with the Xeebra product. Xeebra~\cite{EVS2022Xeebra} is a professional
product for Video Assistant Referee, whose Offside Technology features
are certified by the FIFA Quality Program~\cite{FIFA2019Handbook}.
 With these experiments, our evaluation protocol confirms
the concerns raised~\cite{Theiner2023TVCalib,Chu2022Sports,Claasen2023Video} about the quality of the WC14 ground truth. We obtain different results for the WC14 homographies evaluation than Theiner \etal \cite{Theiner2023TVCalib} which is explained by the fact that we used our annotations. We also establish that
the consolidated annotations provided by Claasen \etal~\cite{Claasen2023Video}
in CARWC are a welcome enhancement in terms of precision. 
\begin{table}
\begin{centering}
\par\end{centering}
\centering{}%
\begin{tabular}{>{\raggedright}m{0.4\columnwidth}|>{\centering}m{0.20\columnwidth}|>{\centering}m{0.25\columnwidth}}
Camera model & $\calibrationMetricWithThreshold 5$ ($\uparrow$) & Reprojection error ($\downarrow$)\tabularnewline
\hline 
Homography (WC14 annotations \cite{Homayounfar2017Sports}) & 67.4 & 3.07\tabularnewline
\hline 
Homography (CARWC annotations \cite{Claasen2023Video}) & 79.1 & 1.79\tabularnewline
\hline 
Pinhole camera parameters with one radial distortion coefficient (\cite{EVS2022Xeebra}) & 92.5 & 1.44\tabularnewline
\end{tabular}\caption{Comparison of camera models on the World Cup 14 dataset.\label{tab:Comparison-of-annotations}}
\end{table}

However, despite visible enhancements provided by the CARWC annotations
to the WC14 dataset, the homography models fail to reproduce the images
of the field markings, while we obtain better results by estimating
camera parameters with Xeebra. This experiment demonstrates that the
qualitative insights shown in Figure~\ref{fig:qualitative-assessment}
are further supported by both the $\calibrationMetricWithThreshold 5$
metric and the reprojection error, comforting the idea that our evaluation
protocol is relevant and fulfills a need for correct evaluation. Moreover,
the middling results of the careful CARWC annotations suggest that
the problem does not lie with the quality of the provided annotations,
but rather with the nature of the annotation. Indeed, we attribute
the better results of the camera parameters of Xeebra to its inclusion
of radial distortion parameters. In an attempt to demonstrate both
our protocol's ability to assess the quality of different camera models
and to further prove our last hypothesis, we evaluated Xeebra's results
on SoccerNet as well.

In the next experiment, as can be seen in Table \ref{tab:Comparison-of-camera-models-SN},
we establish that on the SoccerNet test set, which is a much larger
and thus more diverse and challenging dataset, it is a necessity to
consider the radial distortion. Indeed, when radial distortion is considered in the optimization of the camera parameters, better
results are obtained according to our protocol.

In these experiments, we have shown that our evaluation protocol obtains
results in agreement with qualitative and quantitative evaluations,
and that it allows to evaluate different camera models, which enables
us to support our second contribution, stating that soccer broadcast
cameras are better modeled with radial distortion.

\begin{table}
\centering{}%
\begin{tabular}{cccc}
\hline 
\noalign{\vskip0.1cm}
 & $\calibrationMetricWithThreshold 5$ ($\uparrow$) & $\calibrationMetricWithThreshold 2$ ($\uparrow$) & Reprojection error ($\downarrow$)\tabularnewline[0.1cm]
\hline 
\noalign{\vskip0.1cm}
P & 78.7 & 40.2 & 4.51\tabularnewline
R & 83.1 & 54.3 & 4.01\tabularnewline[0.1cm]
\hline 
\end{tabular}\caption{Comparison of camera models on the SoccerNet-calibration dataset.
P corresponds to the simplified pinhole camera model, and R corresponds
to the simplified pinhole model extended with one radial distortion
coefficient. \label{tab:Comparison-of-camera-models-SN}}
\end{table}

\section{Discussion\label{sec:Discussion}}

\begin{figure*}[t]
\begin{centering}
\includegraphics[width=1\textwidth]{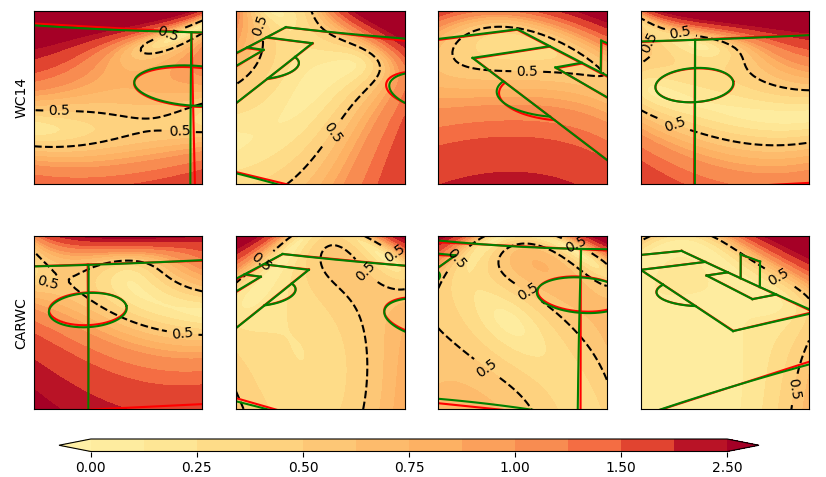}
\par\end{centering}
\caption{Illustration of the disagreement between estimations from two camera models: the ground-truth homography (see the red wireframes) and the richer model combining the pinhole and radial distortion used in Xeebra (see the green wireframes). The small difference between the two reprojected wireframes is misleading, as the superimposition of contour plots on the field shows that some parts of the  sports field are seen over 2.5 meters apart by these two camera models.
We have also plotted the contour line for a difference of 50 cm (see the dashed lines), which is the minimal accuracy standard demanded by the FIFA for Offside Technologies,  to highlight why methods developed with older calibration protocols such as the WC14 or CARWC would fail to meet professional requirements. 
\label{fig:Distance-in-meters}}
\end{figure*}

With the previous results, we have shown that better camera modeling
allows for better precision in the image. In this section, we want
to stress the impact of such improvements on the real-world applications
of camera calibration algorithms.

Considering that many sports field registration methods emphasize
the relevance of their research in applications like player tracking,
it is essential to highlight the pivotal role of selecting an appropriate
camera model for efficient player tracking on a sports field.  In Figure~\ref{fig:Distance-in-meters}, 
we show that, depending on the chosen camera model to link the image
and the physical world, the differences in the estimated 3D positions
can often exceed one meter. Such variations will inevitably impact
the accuracy of player tracking throughout an entire sports game.
Hence, we propose initiating a discussion on optimal modeling strategies
before venturing into applications that may not be practically feasible.
With our newly introduced benchmarking protocol,  we enable a systematic
evaluation that facilitates the exploration and analysis of the best
models in this context.

Still, our protocol presents some shortcomings. The inherent limitations
of our protocol lies in the quantity of field elements present in
the images. Yet, the quality of the evaluation gets better when the
number of elements increases and when the field markings are well
distributed in the image. Indeed, camera models may become over-parameterized
(and be overfitted) in the case of images that show only few field elements.
Another limitation is the fact that it might be possible that very
different camera parameters obtain the same results due to the symmetry
of the sports field template, for instance; the uniqueness of the results
is not guaranteed. However, introducing a means of
managing all kinds of ambiguities, which in practice are rather rare,
would make the evaluation protocol considerably more complex. A possible
yet costly solution to this issue is to acquire real ground truth
for the camera parameters, which requires new datasets and much greater
means to equip the production cameras with sensors.

\section{Conclusion\label{sec:Conclusion}}

In conclusion, this study demonstrates the inadequacy of current
sport field registration benchmarks due to suboptimal ground truth.
Considering the availability of superior camera models, we advocate
abandoning metrics reliant on homographies, and propose a new type
of annotation, combined with proper metrics, leading to the definition
of a camera model-agnostic evaluation protocol. Furthermore, our protocol
 proves that richer camera models, such as the pinhole model augmented
with radial distortion instead of homographies obtain better results
on broadcast cameras, which emphasizes the need to evaluate camera
calibration algorithms independently of the camera model.  Finally,
it should be mentioned that our evaluation methodology applies to
multi-view camera systems, as one can imagine that 3D elements visible
in the calibrated cameras can be used, like any annotated data,
as key points to calibrate other cameras. By doing so, not only have
we have proven that our protocol improves camera calibration in sports,
but also that there are new opportunities that are beyond the reach
of methods solely based on field registration.

\paragraph*{Acknowledgments.}

This work was supported by the Service Public de Wallonie (SPW) Recherche, Belgium, under Grant $\text{N}^{\text{o}}$8573.

\end{document}